\documentclass[sigconf]{acmart}
\usepackage{bm}
\usepackage{multirow}
\usepackage{textcomp}
\AtBeginDocument{%
  \providecommand\BibTeX{{%
    \normalfont B\kern-0.5em{\scshape i\kern-0.25em b}\kern-0.8em\TeX}}}

\if 0
\copyrightyear{2020} 
\acmYear{2020} 
\setcopyright{acmcopyright}\acmConference[KDD '20]{Proceedings of the 26th ACM SIGKDD Conference on Knowledge Discovery and Data Mining}{August 23--27, 2020}{Virtual Event, CA, USA}
\acmBooktitle{Proceedings of the 26th ACM SIGKDD Conference on Knowledge Discovery and Data Mining (KDD '20), August 23--27, 2020, Virtual Event, CA, USA}
\acmPrice{15.00}
\acmDOI{10.1145/3394486.3403319}
\acmISBN{978-1-4503-7998-4/20/08}

\acmSubmissionID{apfp0041}\fi

\begin{document}
\fancyhead{} % To remove the header of all pages
\title[Category-Specific CNN for Visual-aware CTR Prediction at JD.com]{Category-Specific CNN for Visual-aware CTR Prediction at JD.com}

\author{Hu Liu}
\affiliation{\institution{Business Growth BU, JD}}
\email{liuhu1@jd.com}

\author{Jing Lu}
\affiliation{\institution{Business Growth BU, JD}}
\email{lvjing12@jd.com}

\author{Hao Yang}
\affiliation{\institution{Business Growth BU, JD}}
\email{yanghao17@mails.tsinghua.edu.cn}

\author{Xiwei Zhao}
\affiliation{\institution{Business Growth BU, JD}}
\email{zhaoxiwei@jd.com}

\author{Sulong Xu}
\affiliation{\institution{Business Growth BU, JD}}
\email{xusulong@jd.com}

\author{Hao Peng}
\affiliation{\institution{Business Growth BU, JD}}
\email{penghao5@jd.com}

\author{Zehua Zhang}
\affiliation{\institution{Business Growth BU, JD}}
\email{zhangzehua@jd.com}

\author{Wenjie Niu}
\affiliation{\institution{Business Growth BU, JD}}
\email{niuwenjie@jd.com}

\author{Xiaokun Zhu}
\affiliation{\institution{Business Growth BU, JD}}
\email{zhuxiaokun@jd.com}

\author{Yongjun Bao}
\affiliation{\institution{Business Growth BU, JD}}
\email{baoyongjun@jd.com}

\author{Weipeng Yan}
\affiliation{\institution{Business Growth BU, JD}}
\email{paul.yan@jd.com}
\renewcommand{\shortauthors}{Liu and Lu, et al.}

\begin{abstract}
As one of the largest B2C e-commerce platforms in China, JD.com also powers a leading advertising system, serving millions of advertisers with fingertip connection to hundreds of millions of customers.
In our system, as well as most e-commerce scenarios, ads are displayed with images.
This makes visual-aware Click Through Rate (CTR) prediction of crucial importance to both business effectiveness and user experience.
Existing algorithms usually extract visual features using \textit{off-the-shelf} Convolutional Neural Networks (CNNs) 
and \textit{late} fuse the visual and non-visual features for the finally predicted CTR.
Despite being extensively studied, this field still face two key challenges.
First, 
although encouraging progress has been made in offline studies,
applying CNNs in real systems remains non-trivial, due to the strict requirements for efficient end-to-end training and low-latency online serving.
Second, the 
off-the-shelf CNNs and late fusion architectures are suboptimal. 
Specifically, off-the-shelf CNNs were designed for classification thus never take categories as input features.
While in e-commerce, categories are precisely labeled and contain abundant visual priors that will help the visual modeling.
Unaware of the ad category, these CNNs may extract some unnecessary category-unrelated features, wasting CNN's limited expression ability.
To overcome the two challenges, we propose \textbf{Category-specific CNN} (CSCNN) specially for CTR prediction. 
CSCNN \textit{early} incorporates the category knowledge with a light-weighted attention-module on each convolutional layer.
This enables CSCNN to extract expressive category-specific visual patterns that benefit the CTR prediction.
Offline experiments on benchmark and a 10 billion scale real production dataset from JD, together with an Online A/B test show that CSCNN outperforms all compared state-of-the-art algorithms.
We also build a highly efficient infrastructure 
to accomplish 
end-to-end training with CNN on the 10 billion scale real production dataset 
within 24 hours, and meet the low latency requirements of online system (20ms on CPU).
CSCNN is now deployed in the search advertising system of JD, serving the main traffic of hundreds of millions of active users.

\end{abstract}

\maketitle

\section{Introduction}
As one of the largest B2C e-commerce platforms in China,
JD.com also powers a leading advertising system, 
serving millions of advertisers with 
fingertip connection to hundreds of millions of customers.
Every day, customers visit JD,
click ads and leave billions of interaction logs.
These data not only feed the learning system, but also boost technique revolutions that keep lifting both user experience and advertisers' profits on JD.com.

In the commonly used cost-per-click (CPC) advertising system, ads are ranked by effective cost per mile (eCPM), 
the product of bid price given by advertisers and the CTR predicted by the ad system.
Accurate CTR prediction benefits both business effectiveness and user experience. Thus, this topic has attracted widespread interest in both machine learning academia and e-commerce industry.

\if 0
Click through rate(CTR) prediction is a wildly used technique in e-commerce advertising systems. 
In the most common case of cost-per-click(CPC) billing, advertisements are ranked by effective cost per mile(eCPM), which is the product of bid price given by advertisers and CTR given by the system.
Since accurate CTR prediction benefits both business effectiveness and user experience, this topic has attracted widespread interest in both machine learning academia and e-commerce industry.
\fi
Nowadays, most of the ads on e-commerce platforms are displayed with images, since they are more visually appealing and convey more details compared to textual descriptions.
An interesting observation is that many ads get significantly higher CTR by only switching to more attractive images.
This motivates a variety of emerging studies on extracting expressive visual features for CTR prediction \cite{chen2016deep,mo2015image}.
These algorithms adopt various \textit{off-the-shelf} CNNs to extract visual features and fuse them with non-visual features (e.g. \textbf{Category}, user) for the final CTR prediction.
With the additional visual features, these algorithms significantly outperform their non-visual counterparts in \textit{offline}-experiments and generalize well to 
\textit{cold} and \textit{long-tailed} ads.
Although encouraging progress has been made in offline studies, applying CNN in real online advertising systems remains non-trival. 
The offline end-to-end training with CNN must be efficient enough to follow the time-varying online distribution, and the online serving needs to meet the low latency requirements of advertising system.

Furthermore, we notice that visual feature extraction in e-commerce is significantly different from the image classification setting where off-the-shelf CNNs were originally proposed.
In classification, categories are regarded as the target to predict. While in e-commerce, 
categories of ads are clearly labeled, 
which contain abundant visual priors and will intuitively help visual modeling.
Some academic studies have integrated the categorical information by building category-specific projecting matrix on top of the CNN embeddings \cite{he2016sherlock} and by
explicitly decomposing visual features into styles and categories \cite{liu2017deepstyle}.
These studies share a common architecture: the \textbf{late fusion} of visual and categorical knowledge, which however, is sub-optimal for CTR prediction.
Namely, the image embedding modules seldom take advantage of the categorical knowledge. 
Unaware of the ad category, the embedding extracted by these CNNs may contain unnecessary features not related to this category, wasting CNN's limited expression ability.
In contrast, if the ad category is integrated, CNN only needs to focus on the category-specific patterns, which will ease the training process.

To overcome the industrial challenges, we build optimized infrastructure for both 
efficient end-to-end CNN training and low 
latency online servicing.
Base on this efficient infrastructure, we propose \textbf{Category-specific CNN} (CSCNN) specially for the CTR prediction task, to fully utilize the labeled categories in e-commerce.
Our key idea is to incorporate the ad category knowledge into the CNN in an \textbf{early-fusion} manner.
Inspired by the SE-net \cite{hu2018squeeze} and CBAM \cite{woo2018cbam} which model the inter-dependencies between convolutional features with a light-weighted self-attention module, CSCNN further incorporates the ad category knowledge and performs a category-specific feature recalibration, as shown in Fig. \ref{frame}. 
More clearly, we sequentially apply category-specific channel and spatial attention modules to emphasize features that are both important and category related. These expressive visual features contribute to significant performance gain in the CTR prediction problem.

In summary, we make the following contributions:
\begin{itemize}
\item To the best of our knowledge, we are the first to high-light the negative impact of \textbf{late fusion} of visual and non-visual features in visual-aware CTR prediction.
\item We propose CSCNN, a novel visual embedding module specially for CTR prediction. The key idea is to conduct category-specific channel and spatial self-attention to emphasize features that are both important and category related.
 
\item We validate the effectiveness of CSCNN through extensive offline experiments and Online A/B test. We verify that the performance of various self-attention mechanisms and network backbones are consistently improved by plugging CSCNN.
\item We build highly efficient
infrastructure to apply CNN in the real online e-commerce advertising system.
Effective acceleration 
methods are introduced to 
accomplish
the end-to-end training with CNN on the 
10 billion scale
real production dataset within 24 hours, 
and meet the low latency requirements of online system (20ms on CPU).
CSCNN has now been deployed in the search advertising system of JD.com, one of the largest B2C e-commerce platform in China, serving the main traffic of hundreds of millions of active users.
\end{itemize}

\begin{figure}[t]\begin{center}
\includegraphics[width=7.3 cm]{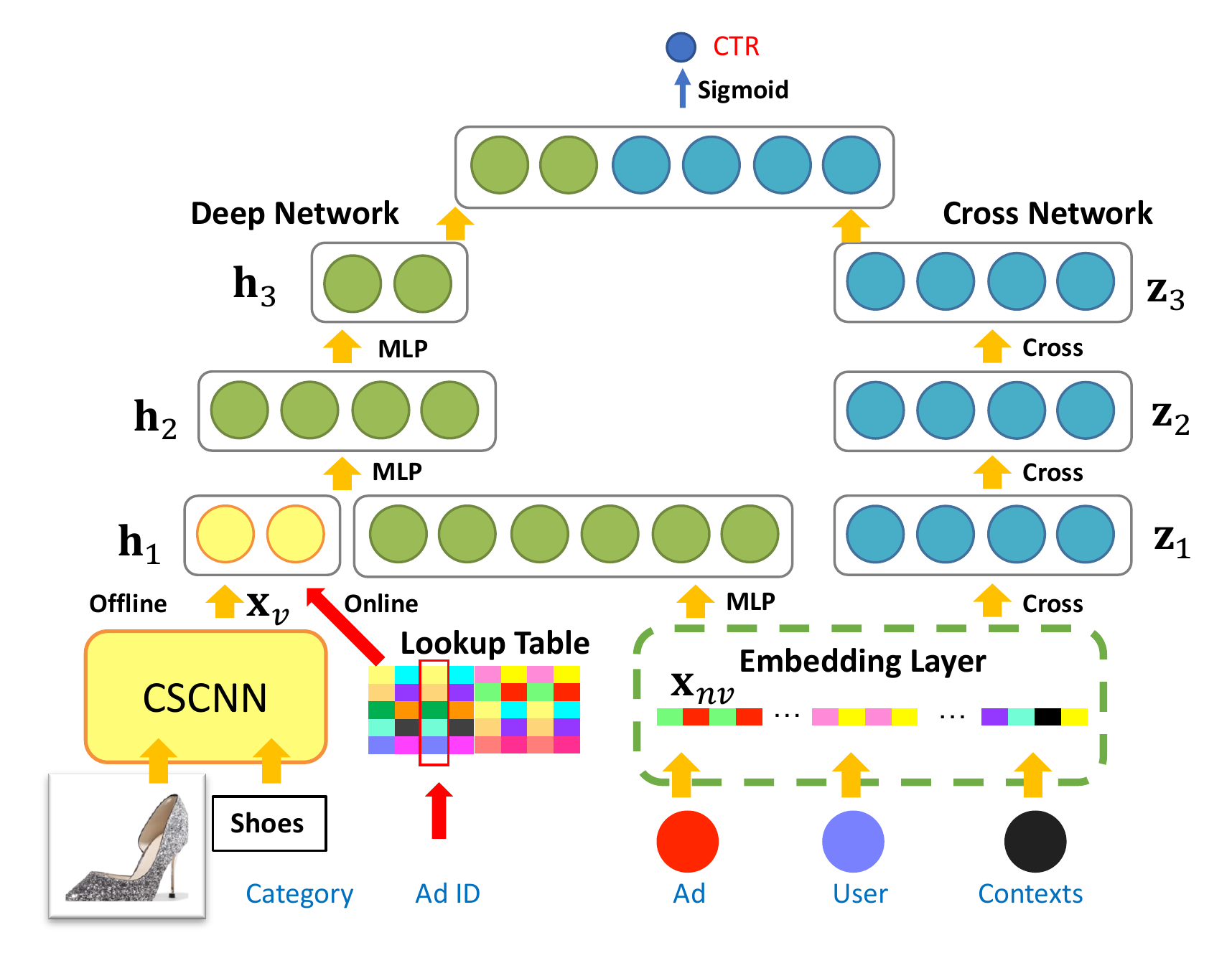}\end{center}
\caption{
The Architecture of our CTR Prediction System. 
Bottom left: the proposed CSCNN, which embeds an ad image together with its category, to a visual feature vector $\mathbf x_{v}\in \mathbb R^{150}$. Note that CSCNN only runs offline. While in the online serving system, to meet the low latency requirement, we use an efficient lookup table instead.
Bottom right: non-visual feature embedding, from (ad, user, contexts) to a non-visual feature vector $\mathbf x_{nv}\in \mathbb R^{380}$.
Top: The main architecture, a modified DCN, which takes both the visual feature $\mathbf x_{v}$ and the non-visual feature $\mathbf x_{nv}$ as inputs.
}\label{arch}
\end{figure}

\section{Related Work}
Our work is closely related to two active research areas: CTR prediction and attention mechanism in CNN.
\subsection{CTR Prediction}
The aim of CTR prediction is to predict the probability that
a user clicks an ad given certain contexts. Accurate CTR prediction benefits both user experience and advertiser's profits thus is of crucial importance to the e-commerce industry.

Pioneer works in CTR or user preference prediction are based on linear regression (LR) \cite{mcmahan2013ad}, 
matrix factorization (MF) \cite{wang2013online} and decision trees \cite{he2014practical}. 
% Pioneer works in CTR or user preference prediction are mainly based on matrix factorization (MF)
% \cite{koren2009matrix,wang2013online,lu2013second}. 
% Later, many algorithms 
% improve the CTR model by learning implicit feature interactions, named the factorization machine  \cite{rendle2010factorization,juan2016field}. 
Recent years have witnessed many successful applications of deep learning in CTR prediction
\cite{cheng2016wide,wang2017deep}.
Early works usually only make use of non-visual features, which however, is insufficient nowadays.  
Currently, most ads are displayed with images which contain plentiful visual details and largely affect users' preference. This motivates many emerging studies on visual aware CTR prediction \cite{chen2016deep,he2016vbpr,kang2017visually,he2016sherlock,liu2017deepstyle,yang2019learning,zhao2019you}.
They first extract visual features using various off-the-shelf CNNs, and then 
fuse the visual features with non-visual features including categories, to build the preference predictor. Unfortunately, 
this late fusion is actually sub-optimal or even a waste in e-commerce scenario, where categories are clearly labeled and contain abundant visual priors that may somehow help visual feature extraction.

In contrast to existing works with late fusion, CSCNN differs fundamentally in early incorporating categorical knowledge into the convolutional layers, %.  Specifically, categorical knowledge is integrated to the self-attention modules after convolution,
allowing easy category-specific inter-channel and inter-spatial dependency learning.

\begin{table}
\caption{Important Notations Used in Section 3}\label{notations}
\begin{center}
\begin{tabular}{ll|ll|ll}
   \hline
$y$&class label				&	$\sigma$ &sigmoid 			&$\hat y$ & predicted CTR	\\
$\mathcal D$&dateset			&	$\mathbb R$ & real number set  	&$d$& feature dimension\\
$\ell$&loss&$\mathbf x$&feature vector		&	  $f$ & prediction function\\
\hline
\end{tabular}
\begin{tabular}{ll|ll}
$l$&layer index&                $\mathbf x_{nv}$& non-visual features\\
$\mathbf h$&hidden layer &	$\mathbf x_{v}$& visual features\\
$\mathbf z$ & cross layer        &	$\mathbf w, \mathbf b$& cross layer parameter \\
$[]$&concatenation		&$d_e$ & embedding dimension\\
$\mathbf x_\text{emb}$ &embedded feature & $E$ & embedding dictionary\\
$v$& vocabulary size	&$\mathbf x_\text{hot}$& one/multi hot coding\\ 
\hline
$m$&ad image&	$k$& ad category \\
$L$	&\# layers 		&	$\mathbf A_s,\mathbf A_s'$		& spatial category prior\\
$C$	& \# channels		&	$\mathbf A_c$	& channel category prior	\\
 $W$ &width 			&$\mathcal K$ & category set\\	 		
 $H$& height &	$\mathbf F$ &  original feature map	\\
 $\mathbf M_s, \mathbf M_c$&attention map 		& $\mathbf F',\mathbf F''$ &  refined feature map\\
 $H',W',C'$ & size of $\mathbf A_s',\mathbf A_c$&	$\odot $ &element-wise product\\ 
  \hline
\end{tabular}\end{center}
\end{table}

\subsection{Attention Mechanism in CNN}
Attention mechanism is an important feature selection approach that
helps CNN to emphasize important parts of feature maps and suppress unimportant ones. 
Spatial attention tells \textit{where} \cite{woo2018cbam} and channel-wise attention tells \textit{what} to  focus on \cite{hu2018squeeze}.  %Usually, the two modules are applied sequentially for a 3D attention map.

In literature, many works have attempted to learn the attention weights from the feature map, termed \textit{self-attention}. 
State-of-the-art algorithms include CBAM \cite{woo2018cbam}, SE \cite{hu2018squeeze} among others \cite{hu2018gather,gao2019global}. 
%Some argues that local attention is too simple to detect inter-channel or spatial relations, so many recent works are proposed to address this issue 
%\cite{zhang2019relation,gao2019global}.
Besides self attention, the attention weights can also be conditioned on external information, for example nature language.  % Generally, the nature language is encoded by an LSTM to get a sequence of feature vectors that will guild the attention module in a CNN based image embedding network. 
  Successful application fields include
search by language \cite{li2017person}, 
image captioning \cite{xu2015show,chen2017sca}
and visual question answering \cite{yang2016stacked}.

Our work is motivated by the attention mechanism. Rather than vision \& language, we design novel architectures to adapt attention mechanism to
address an important but long overlooked issue, the sub-optimal late fusion of vision and non-vision features in CTR prediction.
We combine the advantages of both self-attention and attention conditioned on external information, namely the ad category.
As a result, our image embedding is able to emphasize features that are both important and category related.

\section{The CTR Prediction System in JD.com}
We first review the background of the CTR prediction in Section \ref{pre}. Then we describe the architecture of our CTR prediction system in Section \ref{architecture}.   
We further dig into details of our novel visual modeling module, the Category-Specific CNN in Section \ref{CSCNN}. Finally, we introduce essential accelerating strategies for online deployment in Section \ref{online}. The notations are summarized in Table \ref{notations}.

\subsection{Preliminaries}\label{pre}
In the online advertising industry, when an \textit{ad} is shown to a \textit{user} under some \textit{contexts}, this scenario is counted as an \textit{impression}. The aim of CTR prediction is to predict the probability that a positive feedback, i.e. \textit{click}, takes place in an impression (\textit{ad, user, contexts}).  
Accurate CTR prediction directly benefits both the user experience and business effectiveness, which makes this task of crucial importance to the whole advertising industry. 

CTR prediction is usually formulated as binary classification. Specifically,  the goal is to learn a prediction function $f: \mathbb R^d \to \mathbb R$ from a training set $\mathcal D=\{(\mathbf x_1, y_1),...,(\mathbf x_{|\mathcal D|}, y_{|\mathcal D|})\}$, where $\mathbf x_i \in \mathbb R^d$ is the feature vector of the $i$-th impression and $y_i \in \{0,1\}$ is the class label that denotes whether a click takes place. 

The objective function is defined as the negative log-likelihood:
\begin{equation}
\ell(\mathcal D) = -\frac{1}{|\mathcal D|} \sum_{i=1}^{|\mathcal D|} y_i \log( \hat y_i)+(1-y_i) \log (1-\hat y_i),
\end{equation}
where $\hat y_i$ is the predicted CTR, scaled to $(0,1)$ by sigmoid $\sigma$:
\begin{equation}
\hat y_i=\sigma\left(f(\mathbf x_i)\right).
\end{equation}
\subsection{The Architecture of CTR Prediction System}\label{architecture}
We now describe the architecture of our CTR prediction system that is serving on JD.com. Details are shown in Fig \ref{arch}.

\begin{figure*}[t]\begin{center}
\includegraphics[width=15.6 cm]{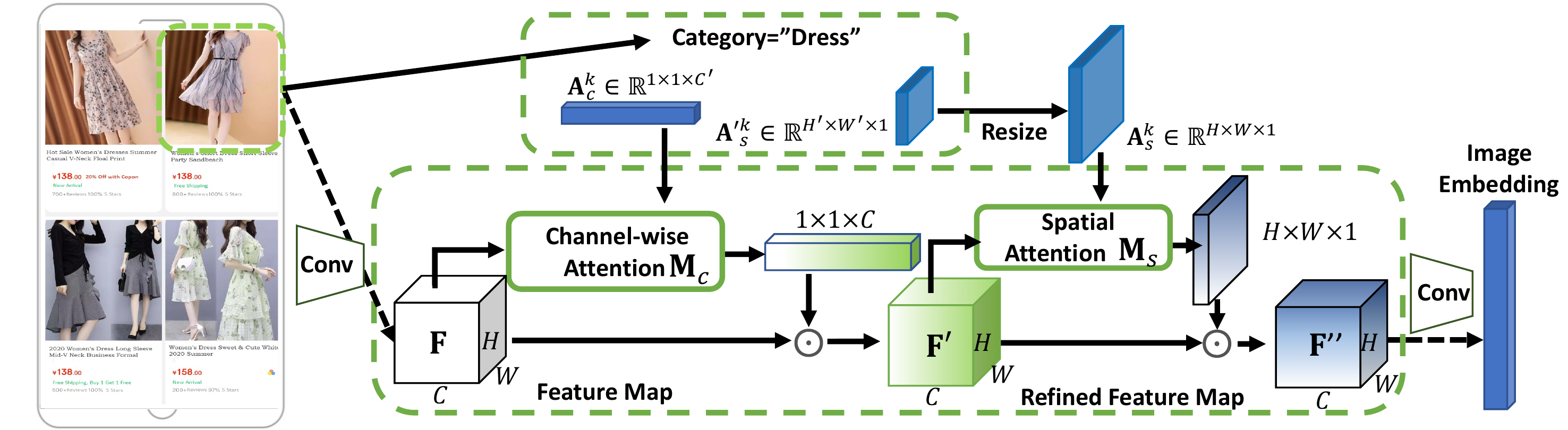}\end{center}
\caption{ 
 Our proposed Category-Specific CNN framework. Note that CSCNN can be added to any single convolutional layer, but for easy illustration, we only show details on single layer.
\textbf{Top}: A map from the category to category prior knowledge that affects channel-wise and spatial attentions.
\textbf{Bottom}:  $\mathbf F$ is the output feature map of the current convolutional layer. Refined by channel-wise and spatial attention sequentially, the new feature map $\mathbf F''$ is used as the input to the next layer.
}\label{frame}
\end{figure*}

\subsubsection{Deep \& Cross Network}
Deep \& Cross network (DCN) \cite{wang2017deep}
has achieved promising performance thanks to the ability to learn effective feature interactions.
Here, we properly modify the DCN to take
two inputs,
a non-visual feature vector $\mathbf x_{nv}\in \mathbb R^{380}$
and
a visual feature vector $\mathbf x_{v}\in \mathbb R^{150}$.

The visual feature is incorporated to the 
\textit{deep} net.
In layer 1, we
 transform the non-visual feature to 1024 dimension and concatenate it with the visual feature, 
\begin{equation}
\mathbf h_{1}=[\mathbf x_{v},\text{ReLU-MLP}(\mathbf  x_{nv})] \in \mathbb R^{150+1024}
\end{equation}
Two deep layers follows,
 \begin{equation}
\mathbf h_{l+1}=\text{ReLU-MLP}(\mathbf  h_l), l\in\{1,2\},\mathbf h_2 \in \mathbb R^{512}, \mathbf h_3 \in \mathbb R^{256}
\end{equation}

The \textit{cross} net is used to process non-visual feature, 
%Following the DCN, a cross layer is defined as
\begin{equation}
\mathbf z_{l+1}=\mathbf z_0 \mathbf z_l ^\top \mathbf w_l+\mathbf b_l +\mathbf z_l,
\end{equation}
where the input $\mathbf z_0=\mathbf x_{nv}$, 
$\mathbf z_l, \mathbf w_l, \mathbf b_l \in \mathbb R^{380}$ for layer $l\in \{0,1,2\}$.

Finally, we combine the outputs for the predicted CTR, 
 \begin{equation}
\hat y = \sigma (\text{ReLU-MLP}[\mathbf h_3, \mathbf z_3])
\end{equation}

\subsubsection{Non-visual Feature Embedding}
We now describe the embedding layer that transforms raw non-visual features of an impression, namely \textit{(ad, user, contexts)}, to the vector $\mathbf x_{nv}$.

We assume that all features come in the categorical form (after preprocessings e.g. binning).
Usually, a categorical feature is encoded in a one-hot / multi-hot vector $\mathbf x_\text{hot} \in \{0,1\}^v$, where $v$ is the vocabulary size of this feature.
We show two examples below:

\textsf{WeekDay=Wed} ~~ ~~~~~$\Longrightarrow$ \textsf{[0,0,0,1,0,0,0]} 

 \textsf{TitleWords=[Summer, Dress]}~~~~~~~$\Longrightarrow$\textsf{[...,0,1,0,...,0,1,0...]}

\noindent Unfortunately, this one/multi-hot coding is not applicable to industrial systems due to the extreme high dimensionality and sparsity. We thus adopt a low dimensional embedding strategy in our system, 
\begin{equation} 
\mathbf x_\text{emb} = E \mathbf x_\text{hot}
\end{equation}
where $E\in \mathbb R^{d_e\times v}$ is the embedding dictionary for this specific feature and $d_e$ is 
the embedding size.
We then concatenate the $\mathbf x_\text{emb}$'s of all features 
to build $\mathbf x_{nv}$.

In practice, our system makes use of 95 non-visual features from users (historical clicks /purchases, location etc.), ads (category, title, \# reviews etc.) and rich contexts (query words, visit time etc.) with 7 billion vocabularies in total. 
Setting $d_e=4$, the total dimension is $95\times4=380$. We will further introduce the features and their statistics in Table \ref{experiment5} Appendix \ref{app_data}.  

\begin{figure*}[t]\begin{center}
\includegraphics[width=12.6 cm]{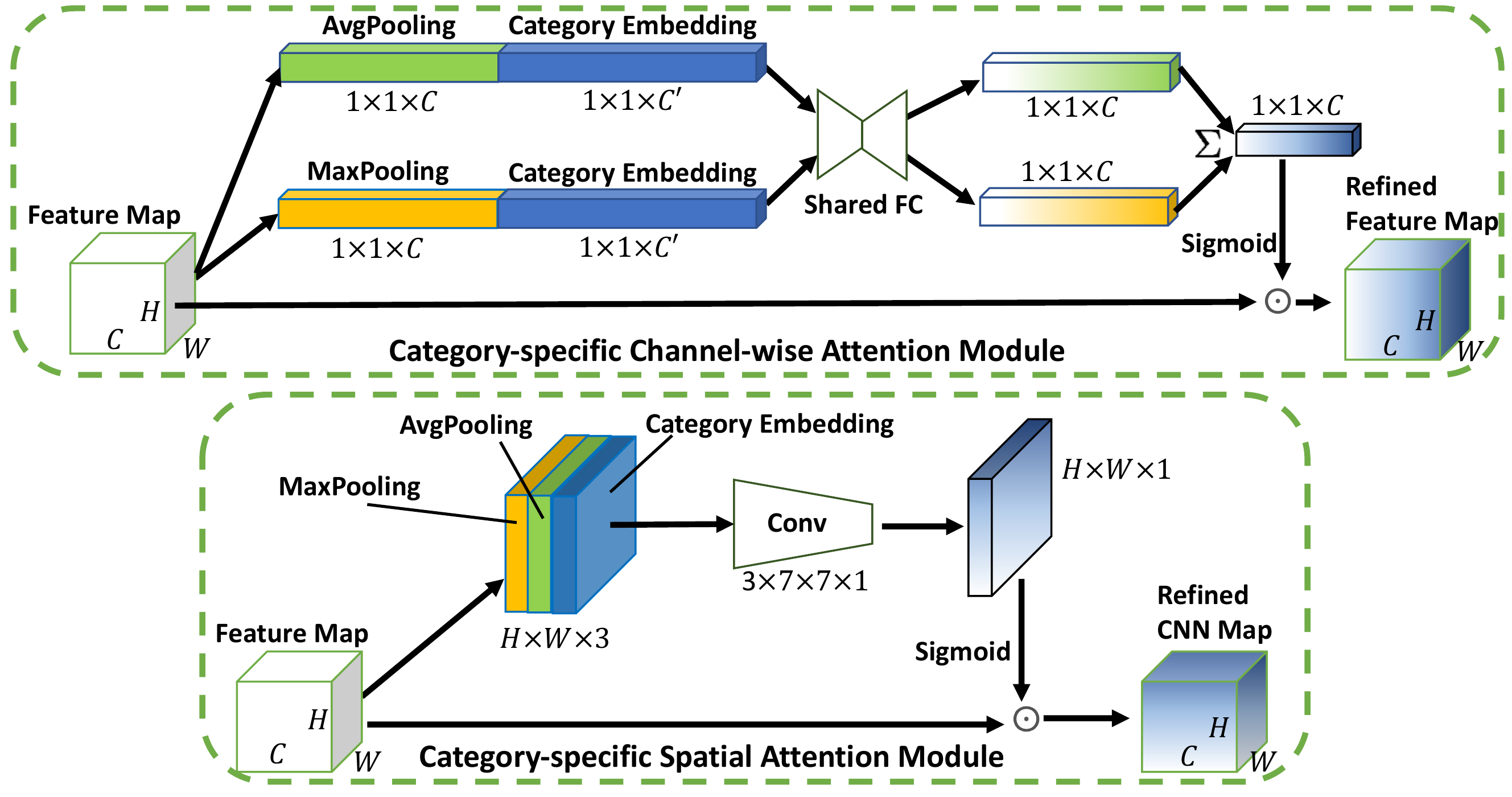}\end{center}
\caption{Modules of our proposed Category-Specific CNN: Channel-wise Attention (top) and Spatial Attention (bottom). }\label{attention2}
\end{figure*}
\subsection{Category-Specific CNN}\label{CSCNN}
%Due to the relevance control of search engine, the search query mostly specifies the category of search results.
Conventional CTR prediction systems mostly 
embed ad images using
\textit{off-the-shelf} CNNs.
We say off-the-shelf since they were 
originally designed for classification, not for CTR prediction.
They regard the image category as the target to predict, not as inputs.
This is actually a huge waste on e-commerce platforms, where categories are precisely labeled and contain plentiful visual prior knowledge that would help visual modeling.   

We address this issue by proposing a novel CNN specifically for CTR prediction,
Category-Specific CNN, that embeds an ad image $m$, together with the ad category $k \in \mathcal K$, to the visual feature $\mathbf x_v$.
Specifically, the category prior knowledge is encoded as category embeddings (trained jointly with the CTR model) and incorporated to the CNN using a conditional attention mechanism.

Theoretically, CSCNN can be adopted to any convoluation layer in any network. In our systems, we plug CSCNN on ResNet18 \cite{he2016deep} and would discuss the adaptability to other networks in ablation studies.  
\subsubsection{Framework on A Single Convolutional Layer}
For each category $k$ and each convolutional layer $l$,
CSCNN learns a tensor  $\mathbf A^k_{c}\in\mathbb R^{1\times 1 \times C'}$  that encodes the impact of category prior knowledge on the channel-wise attention for this layer. We omit the subscript  $l$ for conciseness. The framework is shown in Fig \ref{frame}.

Given an intermediate feature map $\mathbf F \in \mathbb R^{H\times W\times C}$, the output of convolutional layer $l$,  CSCNN first learns a channel attention map $\mathbf M_c\in \mathbb R^{1\times 1\times C}$ conditioned on both the current feature map and the category. Then the channel-wise attention is multiplied to the feature map to acquire a refined feature map $\mathbf F' \in \mathbb R^{H\times W\times C}$, 
\begin{equation}
\mathbf F' =\mathbf M_c(\mathbf F, \mathbf A^k_{c})\odot \mathbf F,
\end{equation}
where $\odot$ denotes the element-wise product with $\mathbf M_c$ broadcasted along spatial dimensions $H \times W$.

Similarly, CSCNN also learns another tensor $\mathbf A^k_{s}\in\mathbb R^{H\times W \times 1}$ that encodes the category prior knowledge for spatial attention $\mathbf M_s\in \mathbb R^{H\times W\times 1}$. These two attention modules are used sequentially to get a 3D refined feature map $\mathbf F'' \in \mathbb R^{H\times W\times C}$,
\begin{equation}
\mathbf F'' =\mathbf M_s(\mathbf F', \mathbf A^k_{s})\odot \mathbf F',
\end{equation}
where spatial attention is broadcasted along the channel dimension before element-wise product. A practical concern is the large number of parameters in $\mathbf A_s^k$, especially on the first a few layers. 
To address this problem, we propose to only learn a much smaller tensor $\mathbf A'^k_{s}\in\mathbb R^{H'\times W' \times 1}$, where $H'\ll H$ and $W'\ll W$, and then resize it to $\mathbf A_s^k$ through linear interpolation. The effect of $H'$ and $W'$ would be discussed with experimental results later. Note that $\mathbf A_s^k$ and $\mathbf A_c^k$ are randomly initialized and learnt during training, no additional category prior knowledge is needed except category id.

After refined by both channel-wise and spatial attention, $\mathbf F''$ is fed to the next layer. Note that  CSCNN could be added to any CNNs, by only replacing the input to the next layer from $\mathbf F$ to $\mathbf F''$.
%Following, we will present the detailed structures inside the two attention modules.
\subsubsection{Category-specific Channel-wise Attention}
Channel-wise attention tells ``what" to focus on. 
In addition to the inter-channel relationship considered previously, we also exploit the relationship between category prior knowledge and features (Fig \ref{attention2}, top).

To gather spatial information, we first squeeze the spatial 
dimension of  $\mathbf F$ through max and average pooling. 
The advantage for adopting both is supported by experiments conducted by the CBAM. 
The two squeezed feature maps are then concatenated with the category prior knowledge $\mathbf A_c^k$ and forwarded through a shared two layer MLP, reducing the dimension from $1\times 1\times (C+C')$ to $1\times 1\times C$. Finally, we merge the two by element-wise summation.
\begin{eqnarray}
\mathbf M_c(\mathbf F,\mathbf A_c^k)= \sigma(\text{MLP}[\text{AvgP}(\mathbf F),\mathbf A_c^k]
+\text{MLP}[\text{MaxP}(\mathbf F),\mathbf A_c^k]),
\end{eqnarray}
\subsubsection{Category-specific Spatial Attention}
Our spatial attention module is illustrated in Fig \ref{attention2} (bottom).
Spatial attention tells where to focus by exploiting the inter-spatial relationship of features. 
Inspired by the CBAM, we first aggregate channel-wise information of feature map $\mathbf F'$ by average pooling and max pooling along the channel dimension. To incorporate the category prior knowledge, these two are then concatenated with $\mathbf A_s^k$ to form an $H\times W \times 3$ dimensional feature map. Finally, this feature map is forwarded through a $7\times 7$ convolutional filter to get the attention weights.
\begin{eqnarray}
\mathbf M_s(\mathbf F',\mathbf A_s^k)= \sigma(\text{Conv}_{7\times 7}(\text{MaxP}(\mathbf F'),\text{AvgP}(\mathbf F'),\mathbf A_s^k)).
\end{eqnarray}

\begin{table}[t]
\caption{\# Parameters and \# GFLOPs of CSCNN and Baselines. We use ResNet18 as the baseline and the backbone network to adopt CBAM and CSCNN modules. Note that there is only 0.03\% addition computation from CBAM to CSCNN. 
} \label{parameters_compare}
\begin{center}
\begin{tabular}{lrr}
\hline
Algorithm & \# params/M & \#GFLOPs\\\hline
Res18& 17.9961   &1.8206\\
Res18 + CBAM& 18.6936   &1.8322\\
Res18 + CSCNN & 21.6791   &1.8329\\
\hline
\end{tabular}\end{center}
%\caption{Dataset Statistics (After Preprocessing). First 3 are from Amazon and the last is from our e-commerce website. Bil. is short for Billion. }\label{dataset}
\end{table}
\subsubsection{Complexity Analysis}
Note that CSCNN is actually a light-weighted module. 
Specifically, we show the number of parameters and giga floating-point operations (GFLOPs) of Baseline, CBAM and our proposed algorithm in Table \ref{parameters_compare}. 

We set $C\in \left\{64, 128,256,512\right\}, C'=20$, and the bottleneck reduction ratio to 4, \# categories $|\mathcal K|=3310$ (real production dataset in Table \ref{industrial}). 
In the ``Shared FC" in each convolutional layer in CBAM, \# parameters is $2*C*C/4$. 
For CSCNN \# parameters in FC and channel category embedding are $C*C/4+(C+C')*C/4+C'*|\mathcal K|$. 
\#params increased compared to CBAM in channel attention for 1 conv layer is 67$\sim$69k.
And, $W'=H'=6$, \#additional params in spatial attention is $W'*H'*|\mathcal K|+6*6\approx$120k.
So, the total params increased is (120k+68k)*16 layers=3.0M. 
The additional params introduced by us are acceptable, and the additional computation is only 0.03\% compared to CBAM.

\begin{figure}[t]\begin{center}
\includegraphics[width=7.8 cm]{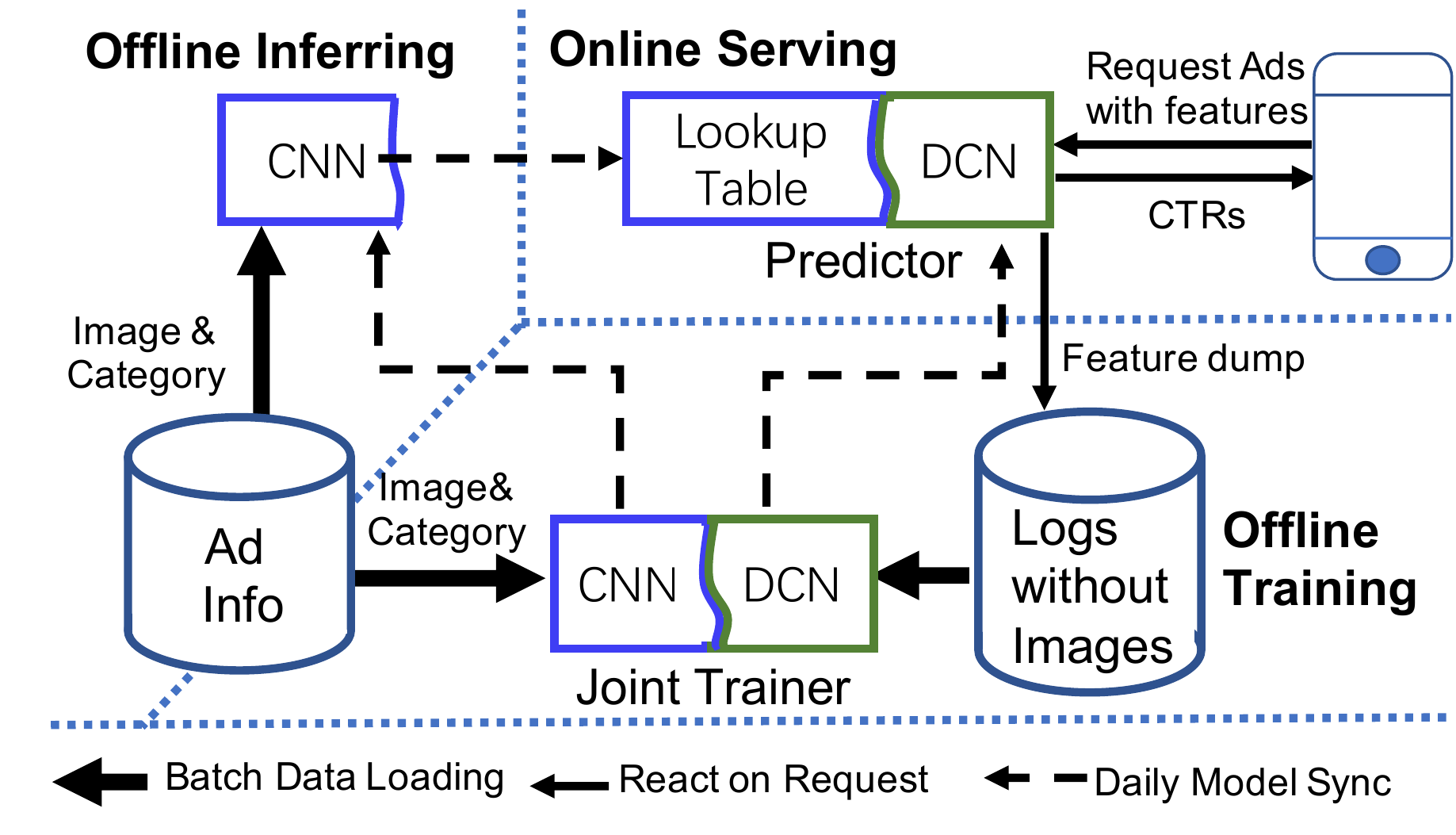}\end{center}
\caption{ 
The architecture of the online model system.
%The architecture of our industrial system. For easier illustration of the visual feature extraction component, we omit other parts of the system (e.g. user features extractor, etc.). The system operates in the following steps: 1. The trainer loads training instances (user-item pairs $u,i$ and the clicked-or-not labels) from the behavior Log database. For each item $i$, the image and category $m_i,k_i$ are loaded from the Item Info database. 2. CSCNN and DCN are trained jointly but serve separately, CSCNN for offline inferring and DCN for online serving. 3. The CSCNN extracts visual features $\Phi(m_i,k_i)\in \mathbb R^d$ of all items. 4. The visual features are loaded to the predictor memory. 5. The predictor receives a request item id $i$. 6. The predictor returns the CTR/ preference score for $i$ predicted from visual features and DCN. And the behavior (clicked-or-not) is recorded to the Log database.
}\label{online_system}
\end{figure}

\subsection{System Deployment}\label{online}
We deploy CSCNN for the search advertising system of JD.com, the largest B2C e-commerce company in China, serving the main traffic of hundreds of millions of active users.
Fig. \ref{online_system} depicts the architecture of our online model system.%, containing offline training, offline inferring and online serving.

\subsubsection{Offline training}
CSCNN is trained jointly with the whole CTR prediction system, on our 
ten-billion scale real production dataset collected in the last 32 days.
In our preliminary investigation, CNN is the key
computational bottleneck during training.
Taking ResNet18 network with input pictures sized 224x224, a single machine with 4 P40 GPUs can only train 177 million images per day.
This means that even only considering CSCNN and with linear speedup in distributed training, we need 226 P40 GPUs to complete the training on the ten-billion impressions within 1 day, which is too expensive. To accelerate, 
we adopt the sampling strategy in \cite{chen2016deep}. 
At most 25 impressions with the same ad are gathered in one batch. The image embedding of one image is conducted only once and broadcasted to multiple impressions in this batch. Now with 28 P40 GPUs, training is can be finished in 1 day.

\if 0
CSCNN and the base DCN \cite{wang2017deep} are jointly trained with the ten-billion scale industrial data collected from the last 32 days.
In our preliminary investigation, we found that CSCNN has serious computational bottlenecks during the offline training.
Taking the medium-sized res18 network as the backbone CNN, input with 224x224 sized pictures, a single machine with 4 P40 GPUs can only train 177 million pictures per day.
This means that even only consider CSCNN and with linear speedup in distributed training, we need 226 P40 GPUs to complete the training of the ten-billion scale industrial dataset within 24 hours, which is too expensive and unnecessary.
We adopt the sampling strategy proposed in \cite{chen2016deep} to accelerate. 
At most 25 training instances with the same item are gathered in one batch. The image embedding of one item is conducted only once and broadcasted to multiple instances.
With this strategy and a parameter server with 28 P40 GPUs, the training is efficient enough for the daily update.
\fi

\subsubsection{Offline inferring:} Images and categories are fed into the well trained CSCNN to inference visual features. Features are made into a lookup table and then loaded in the predictor memory to replace the CSCNN. After dimension reduction and frequency control, a 20 GB lookup table can cover over 90\% of the next day impression.

\subsubsection{Online serving:} Once a request is received, the visual feature is found directly from the lookup table according to ad id. The predictor returns an estimated CTR.
Under the throughput of over 3 million items per second at traffic peak, the tp99 latency of our CPU online serving system is below 20ms.

\begin{table}[t]
\caption{Amazon Benchmark Dataset Statistics.}\label{dataset}
%\caption{Dataset Statistics (After Preprocessing). First 3 are from Amazon and the last is from our e-commerce website. Bil. is short for Billion. }\label{dataset}
\begin{center}
\begin{tabular}{lrrrr}
\hline
Dataset & \#Users &\#Items& \# Interact & \#Category\\
\hline
Fashion&64,583&234,892&513,367&49\\
Women&97,678&347,591&827,678&87\\
Men&34,244&110,636&254,870&62\\
%Industrial& 0.2 bil.& 0.02 bil.& 15 bil.& 3310\\
\hline
\end{tabular}\end{center}
\end{table}

\section{Experimental Results}
We exam the effectiveness of both our proposed visual modeling module CSCNN and the whole CTR prediction system.
The experiments are organized into two groups: 
\begin{itemize}
    \item Ablation studies on CSCNN aims to eliminate the interference from the huge system. 
    We thus test the category-specific attention mechanism by plugging it 
    onto \textit{light-weighted} CNNs with a very \textit{simple} CTR prediction model. We use popular benchmark datasets for 
    repeatability. 
    \item We further exam the performance gain of our CTR prediction system acquired from the novel visual modeling module. Experiments include both off-line evaluations on a ten-billion scale real production dataset collected from ad click logs (Table \ref{industrial}),
    and online A/B testing on the real traffic of hundreds of millions of active users on JD.com.
\end{itemize}

\subsection{Ablation Study Setup}
Our ablation study is conducted on the ``lightest" model. This helps to 
eliminate the interference from the huge CTR prediction system and focus on our proposed category-specific attention mechanism. 

Specifically, our light-weighted CTR model follows the Matrix Factorization (MF) framework, 
VBPR \cite{he2016vbpr}, since it has achieved state-of-the-art performance in comparison with various light models.
The preference score of user $u$ to ad $a$ is predicted as:
\begin{equation}\label{MF} 
\hat y_{u,a}=\alpha+\beta_u+\beta_a+\bm \gamma_u^\top \bm\gamma_a+ \bm\theta_u^\top \Phi(m_a,k_a),
\end{equation}
where $\alpha\in\mathbb R$ is an offset, $\beta_u,\beta_a\in\mathbb R$ are the bias. $\bm \gamma_u \in \mathbb R^{d'}$ and $\bm \gamma_a \in \mathbb R^{d'}$ are the latent features  of $u$ and $a$. $\bm \theta_u \in \mathbb R^{d_v}$ encodes the latent visual preference of $u$. $\Phi$ is the light-weighted CNN. 

Following VBPR \cite{he2016vbpr}, we use CNN-F \cite{chatfield2014return} as the base CNN $\Phi$, which consists of only 5 convolutional layers and 3 fully connected layers. We plug CSCNN onto layers from conv-2 to conv-5.
For comprehensive analysis, we will further test the effect of plugging attention modules on different layers (Figure \ref{parameter}) and our adaptability to other CNN structures (Table \ref{experiment3}) in following sections.

\begin{table*}[t]
\caption{Comparison with State-of-the-arts. For all algorithms, we report the mean over 5 runs with different random parameter initialization and instance permutations. The std $\approx$0.1\%, so the improvement is  extremely statistically \textbf{significant} under unpaired t-test. CSCNN outperforms all due to 3 advantages:  the additional category knowledge, the early fusion of category into CNN, effective structures to learn category-specific inter-channel and inter-spatial dependency. 
 }\label{experiment1}
\begin{center}
\begin{tabular}{ll | r | rr |  rrr|rr}
\hline
&& No Image & \multicolumn{2}{c|}{With Image} & \multicolumn{5}{c}{With Image + Category}\\ \cline{3-10}
\multicolumn{2}{c|}{Datasets}&BPR-MF&VBPR&DVBPR&DVBPR-C&Sherlock&DeepStyle&DVBPR-SCA&Ours\\\hline
\multirow{2}{*}{Fashion} &All&0.6147&0.7557	&0.8011&	0.8022&	0.7640&	0.7530&	0.8032&	\textbf{0.8156}\\
&Cold&0.5334&	0.7476&	0.7712&	0.7703&	0.7427&	0.7465&	0.7694&	\textbf{0.7882}\\
\hline
\multirow{2}{*}{Women} &All&0.6506&0.7238&	0.7624&	0.7645&0.7265&	0.7232&	0.7772&\textbf{0.7931}\\
&Cold&0.5198&0.7086&	0.7078&	0.7099&	0.6945&	0.7120&	0.7273&	\textbf{0.7523}
\\\hline
\multirow{2}{*}{Men} &All&0.6321&0.7079&	0.7491&	0.7549&	0.7239&	0.7279&	0.7547&	\textbf{0.7749}\\
&Cold&0.5331&0.6880&	0.6985&	0.7018&	0.6910&	0.7210&	0.7048&	\textbf{0.7315}\\
\hline
\end{tabular}\end{center}
\end{table*}

\subsection{Benchmark Datasets}
The ablation study is conducted on 3 wildly used  benchmark datasets about products on  \textit{Amazon.com} introduced in \cite{mcauley2015image} \footnote{Many works also use \textit{Tradesy} \cite{he2016vbpr} which is not suitable here due to the absence of category information.}.
We follow the identical category tree preprocessing method as used in  \cite{he2016sherlock}. The dataset statistics after preprocessing are shown in Table \ref{dataset}.

On all 3 datasets, for each user, we randomly withhold one action for validation $\mathcal V_u$, another one for testing $\mathcal T_u$ and all the others for training $\mathcal P_u$,
following the same split as used in \cite{he2016vbpr}. We report test performance of the model with the best AUC on the validation set. 
When testing, we report performance on two sets of items: \textit{All} items,  and \textit{Cold} items with fewer than 5 actions in the training set.

%We also test on our industrial dataset collected from a leading e-commerce website with ten-billion scale interactions. In addition to the user/ item id features used in Eq. (\ref{MF}), we also make use of the abundant features gathered in our e-commerce system, including user features (age, gender, price sensitivity, IP location), item features (category, shop id, price, ratings) and context features (position, time). 

% 这个表格是一个bug。。前三列是从vbpr文章中抄过来的50，后5列是，自己复现的100.不和谐。。正在把前两列改成100重做。如周四前做不成。就删掉前两列，第三列替换成100
\subsection{Evaluation Metrics}
AUC measures the probability that a randomly sampled positive item has higher preference than a sampled negative one,
\begin{equation}
\text{AUC}=\frac{1}{|\mathcal U|}\sum_{u\in\mathcal U} \frac{1}{|\mathcal D_u|} \sum_{(i,j)\in \mathcal D_u} \mathbb I(\hat y_{u,i}>\hat y_{u,j}),
\end{equation}
where $\mathbb I$ is an indicator function. $i,j$ are indexes for ads. $\mathcal D_u=\{(i,j)|(u,i)\in \mathcal T_u \text{ and } (u,j) \notin(\mathcal P_u\cup  \mathcal V_u\cup  \mathcal T_u)\}$. %Namely, given a user $u$, an item is positive if $u$ interacts with it in the test set or is  negative if $u$ never interacts with it in the whole dataset.

In our ablation studies, algorithms are evaluated on AUC which is almost the default off-line evaluation metric in the advertising industry. Empirically, when the CTR prediction model is trained in binary classification, off-line AUC directly reflects the online performance. In JD.com, every 1\textperthousand  increase in off-line AUC brings 6 million dollars lift in the overall advertising income.

%We also conduct extensive online A/B test on our industrial datasets, evaluated by the gain in CTR, CPC and eCPM.
\subsection{Compared Algorithms}
The compared algorithms are
either 
1). representative in covering different levels of available information, or 2). reported to achieve state-of-the-art performance thanks to the effective use of 
category: 

\begin{itemize}
\item \textbf{BPR-MF}: The Bayesian Personalized Ranking (BPR) \cite{rendle2009bpr}, \textit{No} visual features. Only includes the first 4 terms in Eq (\ref{MF}).
\item \textbf{VBPR}: BPR + visual. The \textit{visual} features are extracted from pre-trained and \textit{fixed} CNN \cite{he2016vbpr}.
\item \textbf{DVBPR}: The \textit{visual} feature extractor CNN is trained \textit{end-to-end} together with the whole CTR prediction model \cite{kang2017visually}.
\item \textbf{DVBPR-C}: DVBPR + category. The \textit{Category} information is \textit{late} fused into MF by sharing $\bm \gamma_a$ among items from the same category.
\item \textbf{Sherlock}: DVBPR + category. \textit{Category} is used in the linear transform \textit{after} the visual feature extractor \cite{he2016sherlock}.
\item \textbf{DeepStyle}: \textit{Category} embedding is subtracted from the visual feature to obtain \textit{style information} \cite{liu2017deepstyle}.
\item \textbf{SCA}: This algorithm was originally designed for image captioning \cite{chen2017sca} where features of captioning were used in visual attention. 
To make it a strong baseline in CTR prediction, we slight modify this algorithm by replacing the captioning features to category embedding, so that the category information is early fused into CNN.
\end{itemize}

In literature, some compared algorithms were originally trained with the \textit{pair-wise} loss (see Appendix \ref{appA}), which however, is not suitable for the industrial CTR prediction problem. For CTR prediction, the model should be trained in binary classification mode, or termed \textit{point-wise} loss, so that the scale of $\sigma(f(\mathbf x))$ directly represents CTR.
For fair comparison on the CTR prediction problem, in this ablation study, all algorithms are trained with the point-wise loss. While we also re-do all experiments with the pair-wise loss for consistent comparison with results in literature (Appendix \ref{appA}).

\begin{figure*}[t]\begin{center}\begin{tabular}{rrr}
\includegraphics[width=4.9 cm]{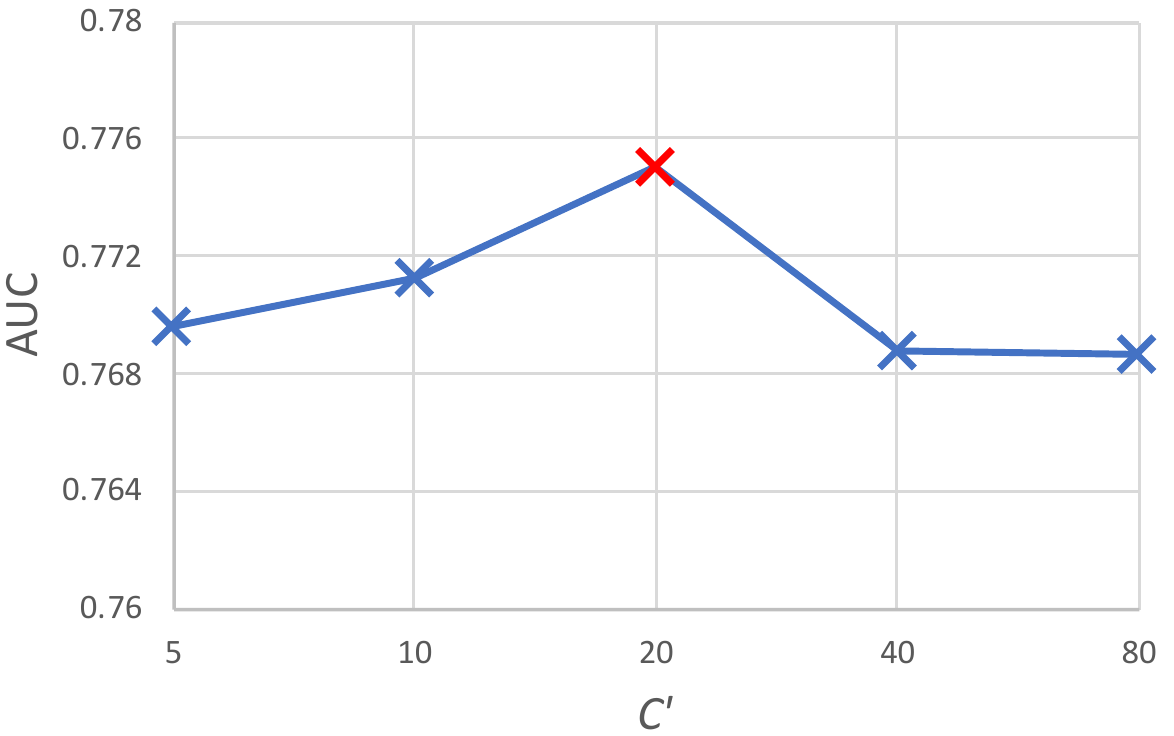}&
\includegraphics[width=4.9 cm]{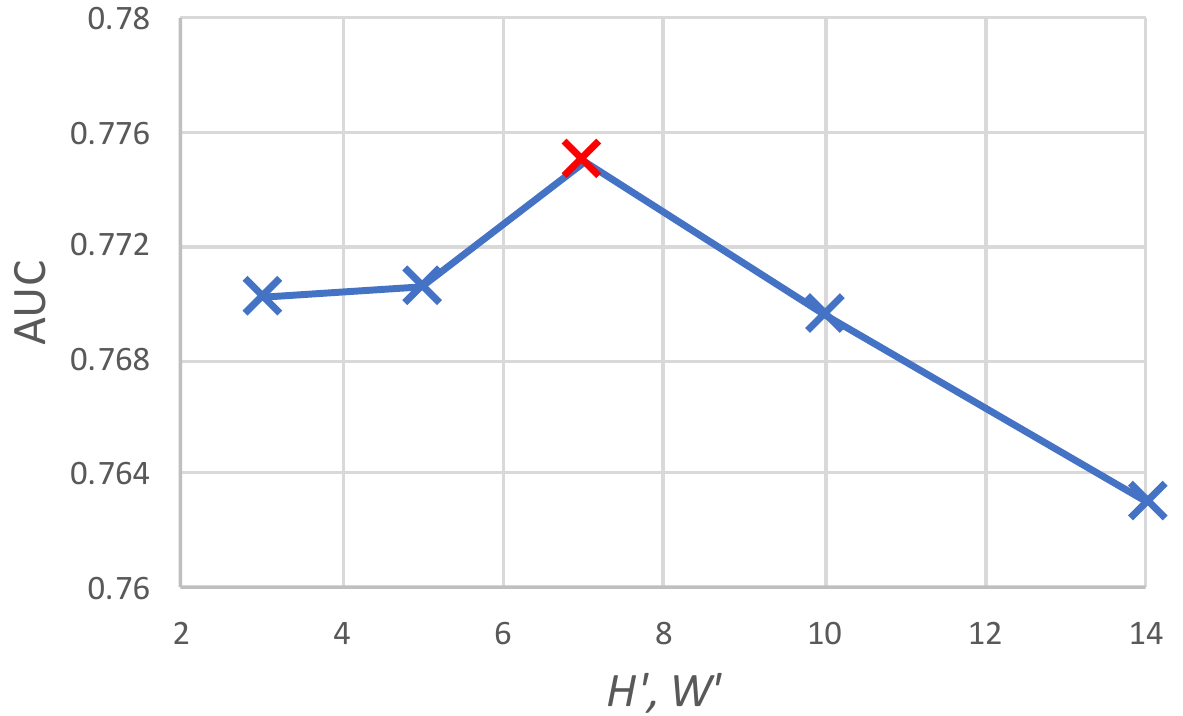}&
\includegraphics[width=4.9 cm]{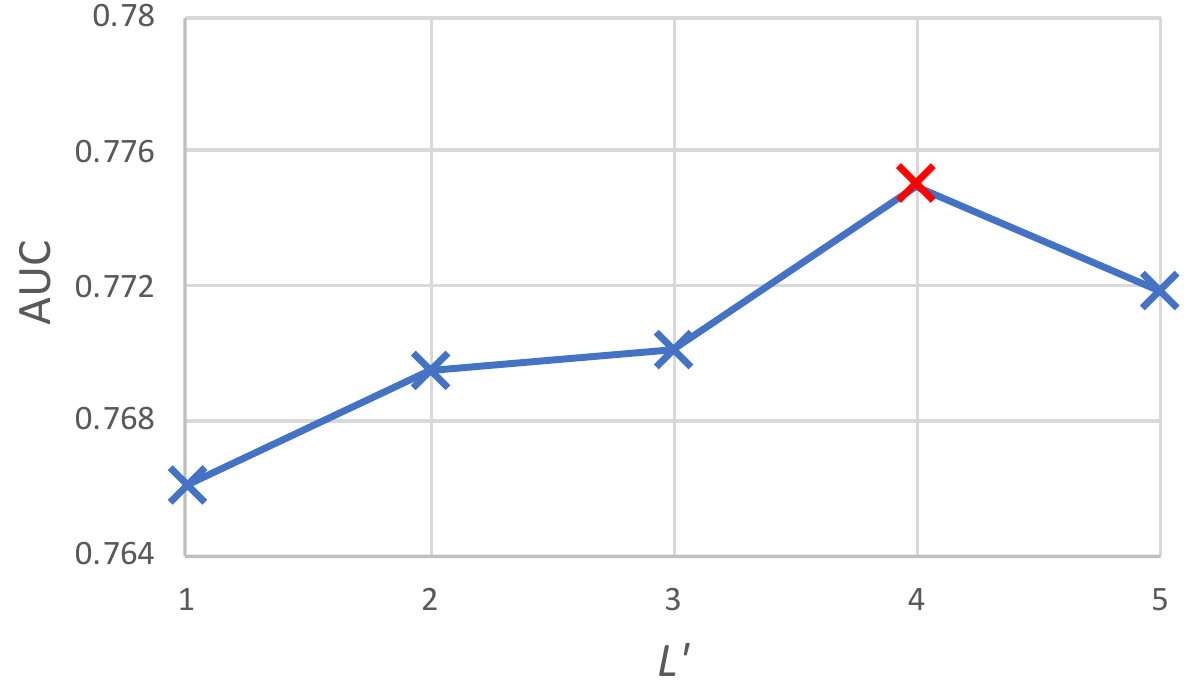}
\end{tabular}\end{center}
\caption{Effects of Hyper-Parameters (Men, all). 
\textbf{$C'$}: The size of category prior $\mathbf A_c^k$ for channel attention.
\textbf{$H',W'$}: The size of category prior $\mathbf A_c^k$ for spatial attention. 
\textbf{$L'$}: Last \# layers to apply CSCNN.
% \textbf{Left \& Middle}: % Effect of $C'$, the size of category prior knowledge $\mathbf A_c^k$ for channel attention.  \textbf{Middle}: Effect of $H'=W'$, the size of  $\mathbf A'^k_{s}$ for spatial attention.
%Before overfitting, increasing $H',W',C'$ benefits AUC. 
%\textbf{Right}: Effect of $L'$. The last $L'$ convolutional layers use CSCNN. More layers use CSCNN, better performance. But using CSCNN on $1_{st}$ layer harms AUC. This is because $1_{st}$ layer learns general, low level features, not category-specific.
}\label{parameter}
\end{figure*}

For fair comparison, all algorithms are implemented in the same environment using Tensor-flow,  mainly based on the source code of DVBPR \footnote{For details, refer to \url{https://github.com/kang205/DVBPR/tree/b91a21103178867fb70c8d2f77afeeb06fefd32c}.}, following their parameter settings, including learning rate, regularization, batch size and latent dimension $d$ etc. Our category prior knowledge dimension is set to $C'=20$, $H'=W'=7$. We will discuss the effects of hyper-parameters in Fig \ref{parameter}. %DVBPR omits $\beta_i+\bm\gamma_u^\top \bm\gamma_i$, we follow. But in DVBPR-C, to incorporate the category information, items from the same category share $\beta_i, \bm \gamma_i$.

\subsection{Comparison with State-of-the-arts}
We aim to show the performance gain from both the incorporating and the 
effective utilization of valuable visual and category information. 
Results are shown in Table \ref{experiment1} \footnote{AUC on the first 3 
columns are from \cite{he2016vbpr}}. %We draw observations:

First, we observe apparent performance gains from additional information along three groups, from ``No Image" to ``With Image", to ``With Image + Category", especially on cold items.
The gain from BPR-MF to group 2  validates the importance of visual features to CTR prediction. 
The gain from VBPR to DVBPR supports significance of end-to-end training, which is one of our main motivation.
And the gain from group 2 to group 3  
validates the importance of category information. 

Second, by further comparing AUC within group 3, where all algorithms make use of category information,  we find that CSCNN  outperforms all. The performance gain lies in the different strategies to use category information.
Specifically, Sherlock and DeepStyle
incorporate the category into a category-specific linear module used at the end of image embedding. 
While in DVBPR-C, items from the same category share the latent feature $\bm \gamma_a$ and $\beta_a$. 
All of them late fuse the category information into the model after visual feature extraction. 
In contrast, CSCNN early incorporates the category prior knowledge into convolutional layers, which enables category-specific inter-channel and inter-spatial dependency learning. 

Third, CSCNN also outperforms DVBPR-SCA, a strong baseline created by us through modifying an image captioning algorithm. Although DVBPR-SCA also early fuses the category priors into convolutional layers through attention mechanism, it lacks effective structures to learn the inter-channel and inter-spatial relationships. While
our effective FC and convolutional structures in $\mathbf M_c$ and $\mathbf M_s$ are able to catch this channel-wise and spatial interdependency.% and thus extract expressive visual features.

\begin{table}[t]
\caption{Adaptability to Various Attention Mechanisms (Amazon Men). 
\textbf{Left}: self attention nets. \textbf{Right}: our modified SE, CBAM-Channel, CBAM-All, by incorporating category prior knowledge $\mathbf A_c^k$ or $\mathbf A_s^k$ to attention using CSCNN. Results with CSCNN (right) significantly outperform original results (left), validating our effectiveness and adaptability.}\label{experiment2}
\begin{center}
\begin{tabular}{l | rr|rr}
\hline
& \multicolumn{2}{c|}{Original} & \multicolumn{2}{c}{+CSCNN}\\ \cline{2-5}
&All&Cold&All&Cold\\\hline
No Attention&0.7491	&0.6985	&--&--\\\hline
SE&0.7500 &	0.6989&	\textbf{0.7673}	&\textbf{0.7153}\\
CBAM-Channel&0.7506&	0.7002&	\textbf{0.7683}&	\textbf{0.7184}\\
CBAM-All&	0.7556&	0.7075	&\textbf{0.7749}&	\textbf{0.7315}\\\hline
\end{tabular}\end{center}
\end{table}

\subsection{Adaptation to Various Attentions}
Our key strategy is to use the category prior knowledge to guide the attention. Theoretically, this strategy could be used to improve any self attention module whose attention are originally learnt only from the feature map.
To validate the adaptability of CSCNN to various attention mechanism, we test it on three popular self attention structures: SE \cite{hu2018squeeze}, CBAM-Channel and CBAM-all \cite{woo2018cbam}.  Their results with self attention is shown in Table \ref{experiment2}, left. 
We slightly modify their attention module using CSCNN, i.e. 
incorporating category prior knowledge $\mathbf A_c^k$ or $\mathbf A_s^k$. Results are in Table \ref{experiment2}, right.

Our proposed attention mechanism with category prior knowledge (right) significantly outperforms their self attention counterparts (left) in all 3 architectures, validating our adaptability 
 to different attention mechanisms. 

\subsection{Adaptability to Various Network Backbones}
As mentioned, our CSCNN can be easily adopted to any network backbones by replacing the input to the next layer from the feature map $\mathbf F$ to the refined feature map $\mathbf F''$. Now we test the adaptability of CSCNN to Inception V1 \cite{szegedy2015going}, results in Table \ref{experiment3}.

CSCNN achieves consistent improvement on CNN-F and Inception V1 over CBAM, which validates our adaptability to different backbones. This improvement also reveals another interesting fact that even for complicated networks (deeper than CNN-F),  there is still big room to improve due to the absent of category-specific prior knowledge. This again supports our main motivation.

\begin{table}[t]
\caption{Adaptability to Different Backbones (Amazon Men). We observe consistent improvement on CNN-F and Inception V1 over the self attention CBAM and no attention base,  validating our adaptability. }\label{experiment3}
\begin{center}
\begin{tabular}{l|l |rr}
\hline
&&CNN-F&Inception\\\hline
\multirow{2}{*}{No Attention} &All& 0.7491	&0.7747\\
&Cold& 0.6985	&0.7259\\
\hline
\multirow{2}{*}{CBAM} &All& 0.7556	&0.7794\\
&Cold& 0.7075	&0.7267
\\\hline
\multirow{2}{*}{CSCNN}&All&	\textbf{0.7749}&\textbf{0.7852}\\
&Cold& \textbf{0.7315}	&\textbf{0.7386}\\
\hline
\end{tabular}\end{center}
\end{table}

\begin{table}[t]
\caption{Real Production Dataset Statistics. Bil, Feats are short for Billion, Features. Besides the features listed, we also do manual feature interaction making the total \# features= 95.}\label{industrial}
\begin{center}

\begin{tabular}{lrll}
\hline
Field& \# Feats & \#Vocab & Feature Example\\
\hline
Ad& 14 & 20M & ad id, category, item price, review\\
User & 6&400 M & user pin, location, price sensitivity\\
Time & 3& 62 & weekday, hour, date\\
Text & 13 & 40M & query, ad title, query title match\\
History&14 & 7 Bil & visited brands, categories, shops\\
\end{tabular}
\begin{tabular}{rrrr}
\hline
\#Users 0.2 bil.&\#Items 0.02 bil.& \# Interact 15 bil. & \#Cate 3310\\
%\#Users &\#Items& \# Interactions & \#Categories\\
% 0.2 bil.& 0.02 bil.& 15 bil.& 3310\\
\hline

\end{tabular}\end{center}
\end{table}

\begin{table}[t]
\caption{Experiments on Real Production Dataset. %\textbf{Top}: offline experiments. \textbf{Bottom}: Online A/B test.
}\label{experiment5}
\begin{center}
\begin{tabular}{l|l}
\hline
Offline & AUC\\\hline
DCN &0.7441\\
DCN + CNN fixed&	0.7463 (+\textbf{0.0022})\\
DCN + CNN finetune&	0.7500 (+\textbf{0.0059})\\
DCN + CBAM finetune&	0.7506 (+\textbf{0.0065})\\
DCN + CSCNN&	0.7527 (+\textbf{0.0086})\\\hline
\end{tabular}

\begin{tabular}{lrrr}
\hline
Online A/B Test &CTR Gain &CPC Gain & eCPM Gain\\\hline
DCN&0&0&0\\
DCN+CSCNN&3.22\%&	-0.62\%&	2.46\%\\\hline
\end{tabular}\end{center}
\end{table}

\subsection{Effects of Hyper-Parameters}
We introduced 3 hyper-parameters in CSCNN: $C'$, the size of category prior $\mathbf A_c^k$ for channel-wise attention; $H'=W'$, the size of category prior $\mathbf A'^k_{s}$ for spatial attention; $L'$, the number of layers with CSCNN module. Namely, we add CSCNN to the last $L'$ convolutional layers of CNN-F. We exam their effects in Fig. \ref{parameter}.

When $H', W', C'$ are small, larger $H', W', C'$ result in higher AUC. This is because larger $\mathbf A_c^k$ and $\mathbf A'^k_{s}$ are able to carry more detailed information about the category, which further supports the significance of exploiting category prior knowledge in building CNN. But too large $H', W', C'$ will suffer from the overfit problem. We find the optimal setting $C'=20$ and $H'=W'=7$. Note that the setting is  restricted to this dataset. For datasets with more categories, we conjecture a larger optimal size of $H', W', C'$.

When $L'\leq 4$, increasing $L'$ benefits. Namely, the more layers that use CSCNN, the better performance. %This again supports the significance of incorporating category into CNN. 
Furthermore, the continuous growth from $L'=1$ to $L'=4$ indicates that the AUC 
 gain from CSCNN at different layers are complementary.
But,  adding CSCNN to \textsf{cov1} harms the feature extractor. This is because \textsf{cov1} learns general, low level features such as line, circle, corner, which never needs attention and are naturally independent to category.

\subsection{Experiments On Real Production Dataset \& Online A/B Testing}
Our real production dataset is collected from ad interaction logs in JD.com. 
We use logs 
in the first 32 days for training and sample 0.5 million interactions from the 33-th day for testing. 
The statistics of our ten-billion scale real production dataset is shown in Table \ref{industrial}.

We present our experimental results in 
Table \ref{experiment5}. 
The performance gain from the fixed CNN validates the importance of visual features.
And the gain from finetuning validates the importance of our end-to-end training system. 
The additional contribution of CBAM demonstrates that emphasizing meaningful features using attention mechanism is beneficial.
Our CSCNN goes one step further by early incorporating the category prior knowledge into the convolutional layers, which enables easier inter-channel and inter-spatial dependency learning. Note that on the real production data, 1\textperthousand increase in off-line AUC is significant and brings 6 million dollars lift in the overall advertising income of JD.com.

From 2019-Feb-19 to 2019-Feb-26, online A/B testing was conducted in the ranking system of JD.
CSCNN contributes 3.22\% CTR (Click Through Rate) and 2.46\% eCPM (Effective Cost Per Mille) gain compared to the previous DCN model online. 
Furthermore, CSCNN reduced CPC (Cost Per Click) by 0.62\%. %Now CSCNN is deployed online and serves the main traffic.

\section{Discussion \& Conclusion}
Apart from the category, what other features could also be adopted to lift the effectiveness of CNN in CTR prediction? 

To meet the low latency requirements of the online systems, CNN must be computed offline. So 
dynamic features including the user and query should not be used.
Price, sales and praise ratio are also inappropriate due to the lack of visual prior.
Visual related item features including brand, shop id and product words could be adopted.
Effectively exploiting them to further lift the effectiveness of CNN in CTR prediction would be a promising direction.

We proposed Category-specific CNN, specially designed for visual-aware CTR prediction in e-commerce. Our early-fusion architecture enables category-specific feature recalibration and emphasizes features that are both important and category related, which contributes to significant performance gain in CTR prediction tasks. With the help of a highly efficient infrastructure, CSCNN has now been deployed in the search advertising system of JD.com, serving the main traffic of hundreds of millions of active users.

%\begin{acks}
%To Robert, for the bagels and explaining CMYK and color spaces.
%\end{acks}

\bibliographystyle{ACM-Reference-Format}
\bibliography{cscnn}

\begin{table*}[h]
\caption{AUC under \textit{Pairwise Loss}, Comparison with State-of-the Art. This table is to demonstrate our consistency to results with pair-wise loss reported in existing works on the same datasets.
All other settings except the loss function are identical to that used in Table \ref{experiment1}. 
Again, our proposed CSCNN framework outperforms all compared algorithms. 
 }\label{experiment1_pairewise}
\begin{center}
\begin{tabular}{ll | r | rr |  rrr|rr}
\hline
&& No Image & \multicolumn{2}{c|}{With Image} & \multicolumn{5}{c}{With Image + Category}\\ \cline{3-10}
\multicolumn{2}{c|}{Datasets}&BPR-MF&VBPR&DVBPR&DVBPR-C&Sherlock&DeepStyle&DVBPR-SCA&Ours\\\hline
\multirow{2}{*}{Fashion} &All&0.6278&0.7479&0.7964&0.7993&0.7558&0.7559&0.7991	&\textbf{0.8157}\\
&Cold&0.5514&0.7319&0.7718&0.7764&0.7359&0.7514&0.7776&\textbf{0.7941}\\
\hline
\multirow{2}{*}{Women} &All&0.6543&0.7081&0.7574&0.7595&0.7185&0.7244&0.7604&\textbf{0.7921}\\
&Cold&0.5196&0.6885&0.7137&0.7191&	0.6919&0.7179&0.7186&\textbf{0.7519}\\\hline
\multirow{2}{*}{Men} &All&0.6450&0.7089&0.7410&0.7483&0.7206&0.7282&0.7515&\textbf{0.7753}\\
&Cold&0.5132&0.6863&0.6923&0.7086&0.6969&0.7247&0.7099&	\textbf{0.7303}\\
\hline
\end{tabular}\end{center}
\end{table*}

\begin{table}[h]
\caption{AUC under \textit{Pairwise Loss}, Adaptability to Various Attention Mechanisms. 
All other settings except the loss function are identical to that in Table \ref{experiment2}.
}\label{experiment2_pairwise}
\begin{center}
\begin{tabular}{l | rr|rr}
\hline
& \multicolumn{2}{c|}{Original} & \multicolumn{2}{c}{+CSCNN}\\ \cline{2-5}
&All&Cold&All&Cold\\\hline
No Attention&0.7410 &0.6923&--&--\\\hline
SE&0.7517&	0.7094&	\textbf{0.7637}&\textbf{0.7219}\\
CBAM-Channel&0.7541&0.7094&\textbf{0.7650}&\textbf{0.7219}\\
CBAM-All&0.7559&0.7081&\textbf{0.7753}&\textbf{0.7303}\\\hline
\end{tabular}\end{center}
\end{table}

\begin{table}[h]
\caption{AUC under \textit{Pairwise Loss}, Adaptability to Different Backbones (Amazon Men). 
All other settings except the loss function are identical to that in Table \ref{experiment3}.
}\label{experiment3_pairwise}
\begin{center}
\begin{tabular}{l|l |rr}
\hline
&&CNN-F&Inception\\\hline
\multirow{2}{*}{No Attention} &All&0.7410&0.7706\\
&Cold&0.6923&0.7314\\
\hline
\multirow{2}{*}{CBAM} &All&0.7559&0.7751\\
&Cold&0.7081&0.7344
\\\hline
\multirow{2}{*}{CSCNN}&All&\textbf{0.7753}&\textbf{0.7824}\\
&Cold&\textbf{0.7303}&	\textbf{0.7461}\\
\hline
\end{tabular}\end{center}
\end{table}
\newpage
\appendix
\section{Additional Experimental Results}\label{appA}
In this appendix, we show many additional experimental results that are essential in supporting our claims and consistency with results reported in related works.

In all our previous ablation studies (Table \ref{experiment1}, \ref{experiment2} and \ref{experiment3}) and training for our online serving systems, we used \textbf{point-wise} loss. Namely, each impression is used as an independent training instance for binary classification. 
Although this is the default setting for the CTR prediction problem, we find that some existing works, which are also tested on \textit{Amazon}, use the \textbf{pair-wise} loss (e.g. \cite{rendle2009bpr,he2016vbpr,kang2017visually}). 

Specifically, a model with pair-wise loss is trained on a dataset of triplets $(u,i,j)\in \mathcal D$, where $(u,i,j)$ indicates that user $u$ prefers ad $i$ to ad $j$.  %So the preference score $x_{u,i}$ should be larger than $x_{u,j}$. 
Following Bayesian Personalized Ranking (BPR), their objective function is defined as
\begin{equation} 
\max \sum_{(u,i,j)\in \mathcal D} \ln \sigma (\hat y_{u,i}-\hat y_{u,j}) -\lambda R,
\end{equation}
where $R$ is the regularization on all parameters.

To make direct comparison with existing results of pair-wise loss on the same dataset, here we re-do our ablation studies using \textbf{pair-wise loss}, with all other settings identical to 
that in Table \ref{experiment1}, \ref{experiment2} and \ref{experiment3}. 
The results are shown in Table \ref{experiment1_pairewise}, \ref{experiment2_pairwise} and \ref{experiment3_pairwise}.

From these results, we can draw several observations. First, comparing our results in Table \ref{experiment1_pairewise}, \ref{experiment2_pairwise} and \ref{experiment3_pairwise} with that reported in related works, we confirm that our reimplementation of pair-wise loss and compared algorithms is comparable or sometimes better than the original literature. This validates our consistency and repeatability. 
Second, our CSCNN framework outperforms all compared algorithms in Table \ref{experiment1_pairewise}, \ref{experiment2_pairwise} and \ref{experiment3_pairwise}, validating the advantages of the CSCNN framework. This superiority indicates that all our previous claims on point-wise loss still hold on pair-wise loss based models. Namely, plugging the CSCNN framework on various self-attention mechanisms and network backbones brings consistently improvement. Third, when comparing the performance across pair-wise and point-wise losses ( Table \ref{experiment1_pairewise} vs. \ref{experiment1}, Table \ref{experiment2_pairwise} vs. \ref{experiment2} and Table \ref{experiment3_pairwise} vs. \ref{experiment3}), we find that neither of the two losses enjoys absolute superiority over the other one in terms of AUC. However, the metric AUC only measures the relative preference of $\hat y_{u,i}$ compared to $\hat y_{u,j}$, not the absolute scale. While in practical advertising industry,
point-wise loss is usually preferred since the scale of $\sigma(f(\mathbf x))$ trained in binary classification directly reflects the click through rate.

\section{Statistics of Our Real Production Datasets}\label{app_data}
\begin{figure*}[t]\begin{center}\begin{tabular}{rrr}
\includegraphics[width=5 cm]{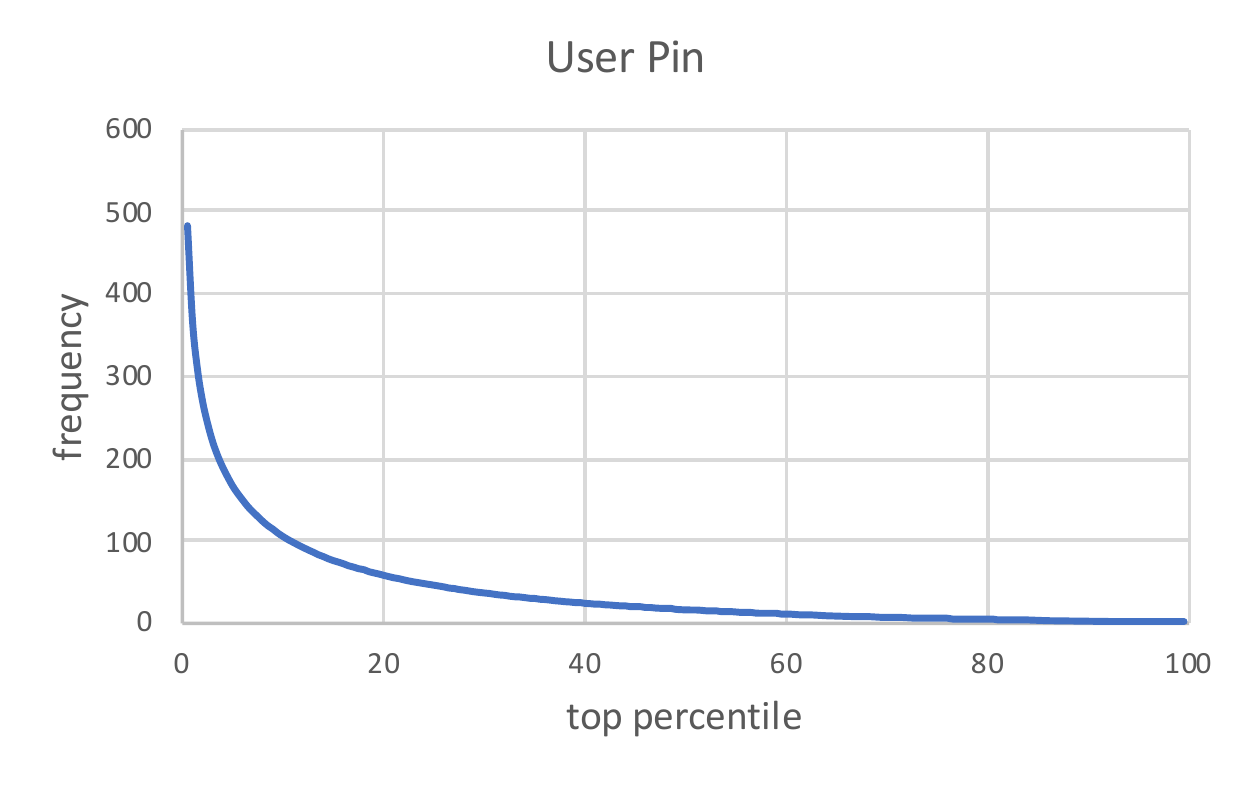}&
\includegraphics[width=5 cm]{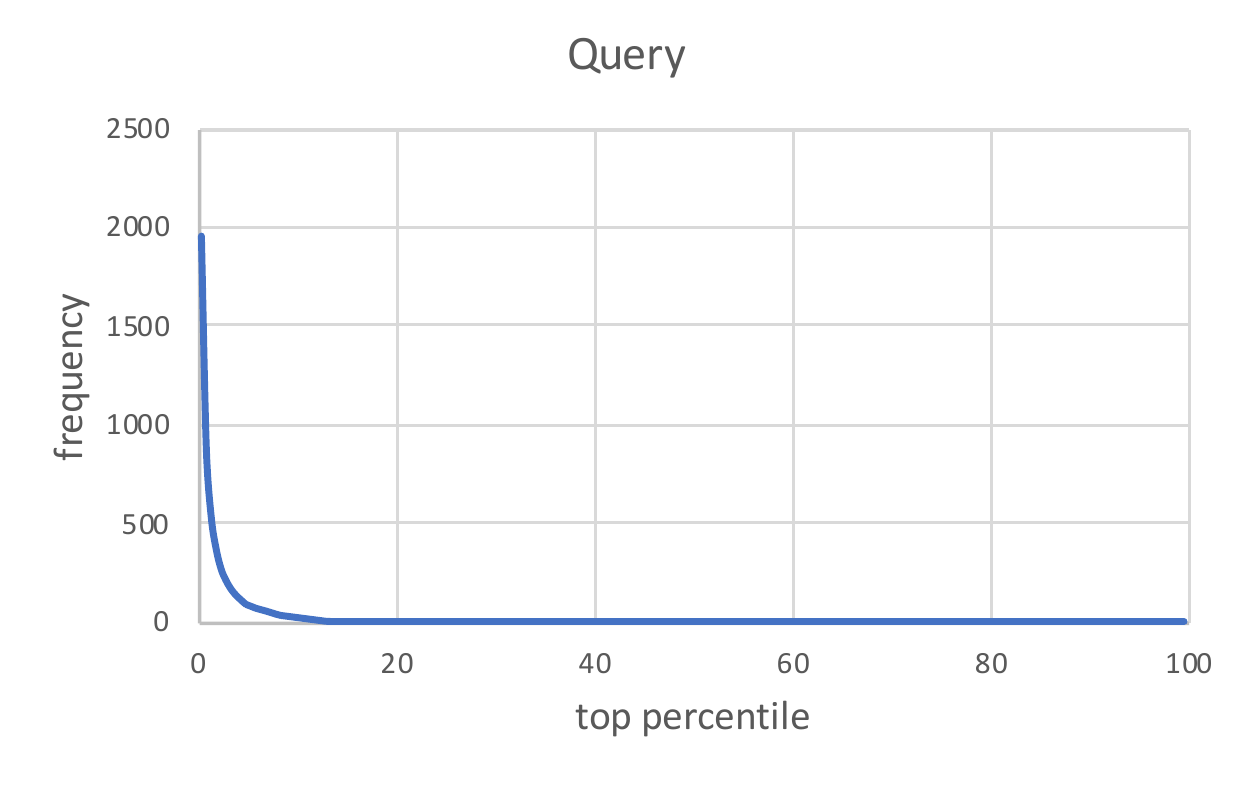}&
\includegraphics[width=5 cm]{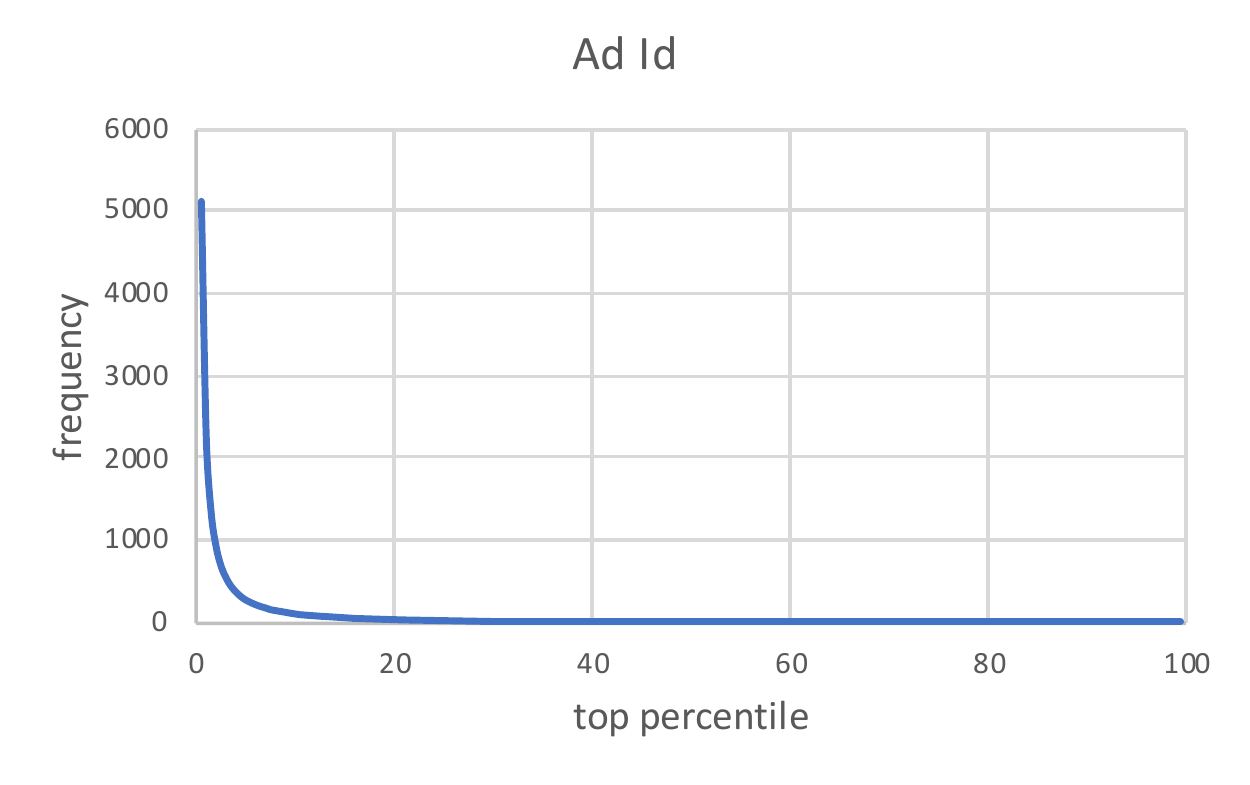}
\end{tabular}
\begin{tabular}{rrr}
\includegraphics[width=5 cm]{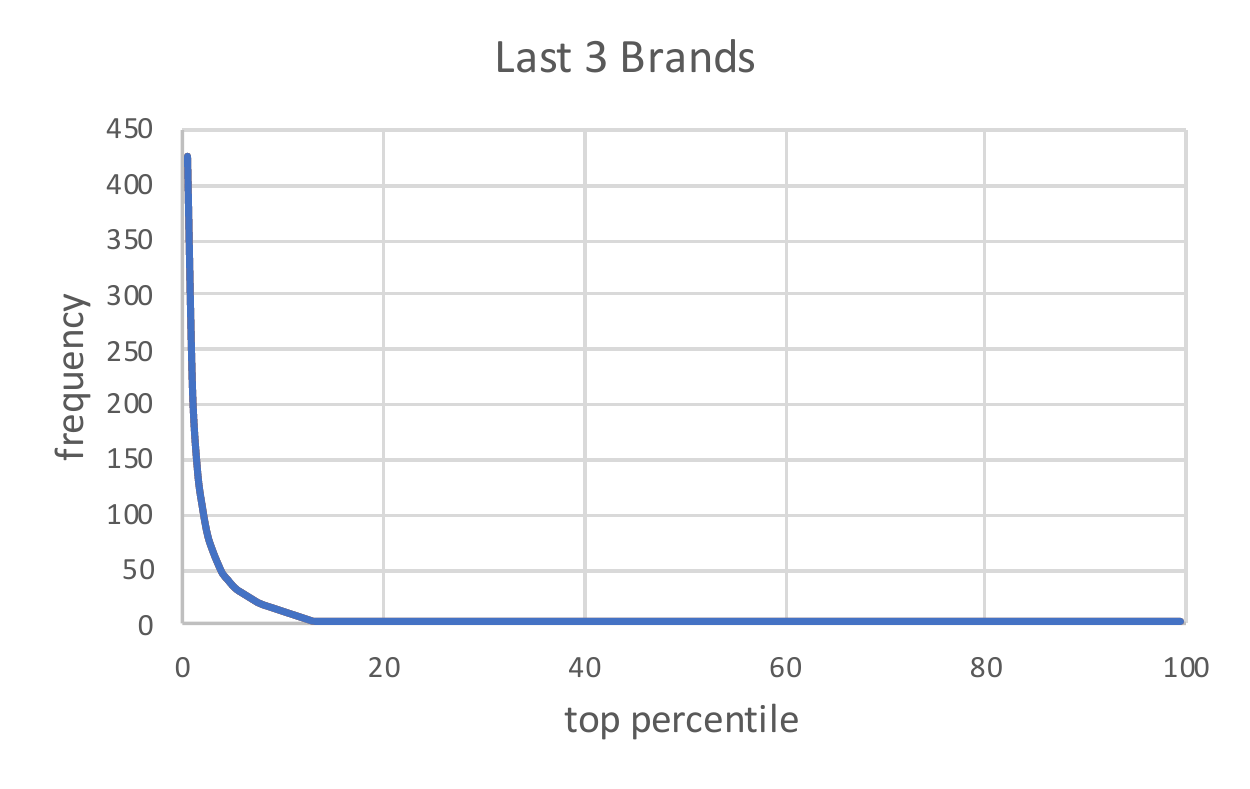}&
\includegraphics[width=5 cm]{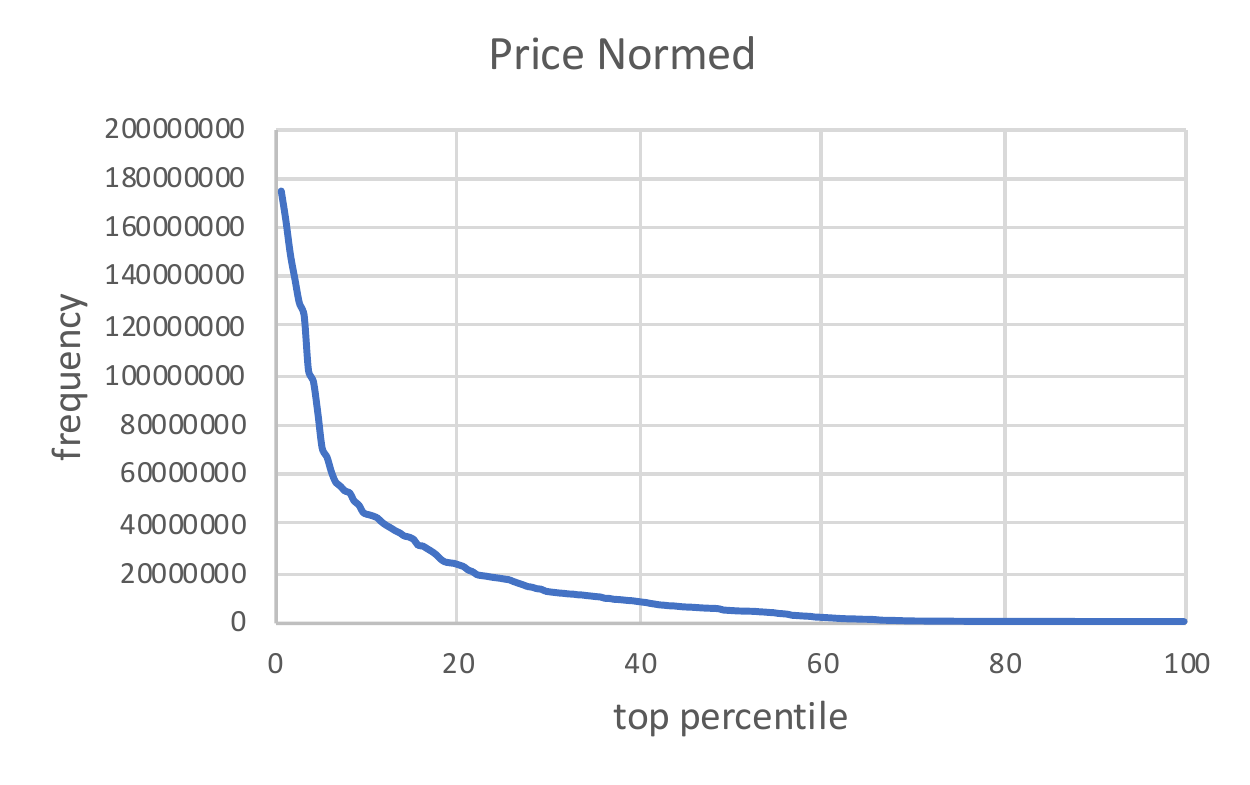}&
\includegraphics[width=5 cm]{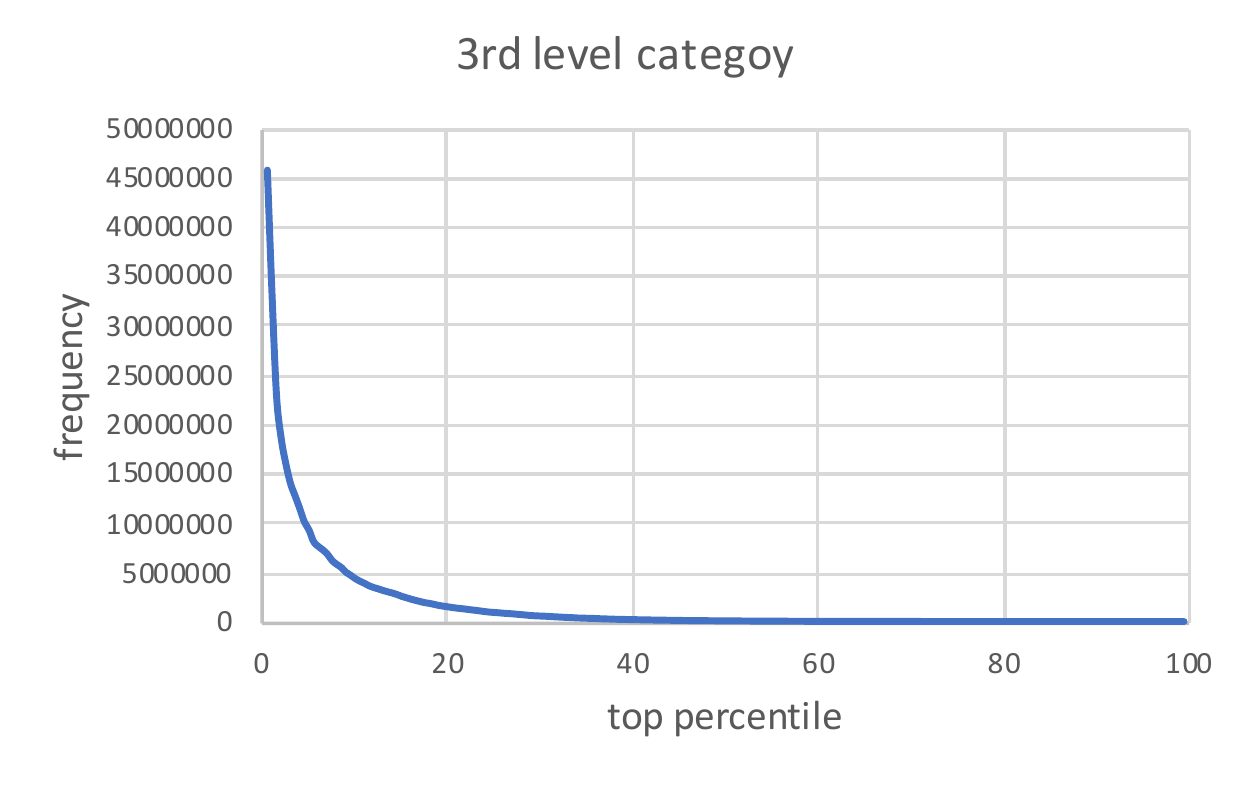}
\end{tabular}\end{center}
\caption{Feature statistics from the search advertising system of JD.com from 20200106 to 20200206 (32 days). For each feature, values are sorted in descending order of frequency.}
\label{featurestats}
\end{figure*}

In this appendix, we show some specific statistics of our real production datasets, see Figure \ref{featurestats}.  
Most of the features are extremely sparse, e.g. 80\% of the queries have appeared less than 6 times in the training set, and follow the long tail distribution.
User pin and price are relatively evenly distributed, but still 10\% of the features cover 50\% of the training set.
As claimed in earlier studies \cite{yang2019learning}, visual features contributes more when other features are extremely sparse.
These statistics illustrate the difficulties in modeling on the real production dataset and the contribution of our methods.

\end{document}